\pdfoutput=1
\documentclass[11pt]{article}

\usepackage{EMNLP2023}

\usepackage{times}
\usepackage{latexsym}

\usepackage[T1]{fontenc}
\usepackage[utf8]{inputenc}

\usepackage{microtype}

\usepackage{inconsolata}

\usepackage{amsmath}
\usepackage{amssymb}

\usepackage{graphicx}

\usepackage{makecell}
\usepackage{multirow}

\usepackage{booktabs}

\usepackage{cleveref}

\title{Baby Llama: knowledge distillation from an ensemble of teachers trained on a small dataset with no performance penalty}

\author{Inar Timiryasov\thanks{~~Equal contributions} \\
  Niels Bohr Institute, \\
  University of Copenhagen, \\
  Blegdamsvej 17, DK-2010, \\
  Copenhagen, Denmark \\
  \texttt{inar.timiryasov@nbi.ku.dk} \\\And
  Jean-Loup Tastet\footnotemark[1] \\
  Departamento de Física Teórica and \\
  Instituto de Física Teórica UAM/CSIC, \\
  Universidad Autónoma de Madrid, \\
  Cantoblanco, 28049, Madrid, Spain \\
  \texttt{jean-loup.tastet@uam.es} \\}

\begin{document}
\maketitle
\begin{abstract}
We present our submission\footnote{\href{https://huggingface.co/timinar/baby-llama-58m}{\texttt{https://huggingface.co/timinar/baby-llama-58m}} for the checkpoint; the training code is available at \href{https://github.com/timinar/BabyLlama}{\texttt{https://github.com/timinar/BabyLlama}}\,.} to the BabyLM challenge, whose goal was to improve the sample efficiency of language models.
We trained an ensemble consisting of a GPT-2 and small LLaMA models on the developmentally-plausible, 10M-word BabyLM dataset, then distilled it into a small, 58M-parameter LLaMA model, which exceeds in performance both of its teachers as well as a similar model trained without distillation.
This suggests that distillation can not only retain the full performance of the teacher model when the latter is trained on a sufficiently small dataset; it can exceed it, and lead to significantly better performance than direct training.
\end{abstract}

\section{Introduction}

Today's state-of-the-art language models are typically trained on the order of a trillion tokens.
\citet{hoffmann2022training} have observed that in order to train a model in a compute-optimal way, the number of parameters and dataset size should follow a linear relation: the so-called Chinchilla scaling law, with an optimal ratio of about 20~tokens per model parameter.
For models larger than $\sim 10^{11}$ parameters, this implies that the currently-available amount of training data ($\sim 10^{12}$ tokens) already constitutes a bottleneck, that prevents scaling up those models in a compute-optimal way.

A trillion tokens is already at least 4 orders of magnitude larger than the estimated number of words\footnote{Extrapolating from \citet{mapping-language-development}.} ($\lesssim 10^8$) to which a typical 13-year-old child has been exposed. This suggests that current language models are significantly less \textit{sample-efficient} than human beings.

Furthermore, the trend of scaling up models to improve their performance may limit their usage in embedded systems, personal devices, and other end-user technologies, as well as in specialized applications where domain-specific training material is scarce.
\citet{taylor2022galactica} have shown that training models on higher-quality data can improve performance; however, the quantity of such high-quality data is limited, and often represents only a small fraction of the corpus.

This makes a strong case for trying to increase the sample efficiency of current models and training algorithms.
In this context, the BabyLM challenge \citep{warstadt2023papers} has invited researchers to investigate ways of improving the sample efficiency of small-scale language models, by restricting the training set to a \textit{developmentally plausible} corpus, consisting mostly of transcribed speech of either 10M (\texttt{strict-small} track) or 100M words (\texttt{strict} and \texttt{loose} tracks).

The present paper describes our submission to the \texttt{strict-small} track of the BabyLM challenge. As such, it focuses on the 10M-word dataset.
Our proposed solution consists in distilling an ensemble of two larger ``teacher'' models, of different architectures (GPT-2 and LLaMA), into a smaller ``student'' LLaMA model. We show that this approach produces a model whose performance largely matches, and often exceeds, that of both teachers.

We introduce Baby Llama in \cref{sec:model}, describe the dataset in \cref{sec:dataset}, discuss the model performance in \cref{sec:performance}, and finally conclude in \cref{sec:conclusion}.
The full numerical results of the evals are listed in \cref{app:evals}, and in \cref{app:attempts} we briefly discuss a number of experiments (including some negative results) that we eventually chose not to include into the final model.

\section{Pretraining using distillation}
\label{sec:model}

Knowledge distillation~\citep{Bucila2006ModelC, 2015arXiv150302531H} is a technique that consists in training a (usually smaller) student model to reproduce the behaviour of one or more teacher models. This method has been successfully applied to large language models, e.g.\ in~\citet{distilbert}.

In our submission to the \texttt{strict-small} track of the BabyLM challenge, we address the sample efficiency problem by distilling an ensemble of larger pre-trained teacher models into a smaller student model. Specifically, we train an ensemble consisting of GPT-2 \citep{gpt2} and a small LLaMA model \citep{llama} on the 10M-word BabyLM dataset, and then distill this ensemble into a smaller, 58M-parameter LLaMA model.
Despite its reduced size, our distilled LLaMA model not only retains the performance of the larger models, but also exceeds it. This shows that distillation can be a powerful tool to enhance sample efficiency when training on smaller datasets.

The distillation process involves guiding the training of the student model using the output of the teacher models. This output, also known as soft targets, is obtained by applying a temperature scaling factor to the teacher's output logits. The student model is then trained to approximate these soft targets (with the same temperature) in addition to the original hard targets, resulting in a model that generalizes better and therefore performs better on unseen data.

The loss function consists of a weighted sum of the original hard target loss (cross-entropy with the true labels) and the distillation loss (Kullback-Leibler divergence with the teacher's soft targets). Formally, it can be expressed as:
\begin{equation}
L = \alpha L_{CE} + (1 - \alpha) L_{KL}
\label{eq:distil-loss}
\end{equation}
where $\alpha$ is the weight factor, $L_{CE}$ is the original cross-entropy loss, and $L_{KL}$ is the Kullback-Leibler divergence.

The teacher models used for the distillation are newly-trained instances of GPT-2 and LLaMA. 
The GPT-2 model has 24 layers, 16 attention heads, an embedding dimension of 1536, intermediate size of 6144, and maximum sequence length of 128, resulting in 705M parameters. It was trained for 6 epochs with a batch size of 256 and maximum learning rate\footnote{We trained all three models using a cosine learning rate schedule with a warm-up of 200 steps.} of $2.5\cdot 10^{-4}$.
The LLaMA model has 24 layers, 8 attention heads, a hidden size of 1024, intermediate size of 3072, and maximum sequence length of 256,  resulting in 360M parameters. It was trained for 4 epochs with a batch size of 128 and maximum learning rate of $3\cdot 10^{-4}$.
Both teacher models are pretrained exclusively on the 10M-word BabyLM dataset. We use the same tokenizer for both the teacher and student models, with a vocabulary size of 16000; the tokenizer is trained exclusively on the \emph{training} split.

For the student model, we chose a smaller version of the LLaMA model with only 16 layers, 8 attention heads, a hidden size of 512 and an intermediate size of 1024, resulting in 58M parameters.
This choice was mainly motivated by the requirement of being able to fine-tune the model with our limited computational resources\footnote{It would be interesting to see if a bigger model --- possibly larger than the teachers --- can be successfully pretrained in the same way.}
for the various benchmark tasks that require fine-tuning.
The distillation process is carried out using a batch size of 32 and a maximum learning rate of $3\cdot 10^{-4}$. The loss function \eqref{eq:distil-loss} is used throughout the entire training, i.e.\ the student model is \emph{not} trained conventionally before the distillation. The training lasts for 6 epochs. The temperature was set to 2 and $\alpha = 0.5$.
We have tried various combinations of 2, 4, and 6 teacher models, with the best results being achieved using two teachers.

We observed that the eval loss did not correlate sufficiently well with the benchmarks to be able to use it as a proxy for the final model performance. 
Therefore, given the limited time and resources, we were not able to perform a systematic hyperparameter search.

The trained model can be downloaded from the HuggingFace repository \url{https://huggingface.co/timinar/baby-llama-58m}\,.
When implementing the distillation loss, we largely followed repository \href{https://github.com/philschmid/knowledge-distillation-transformers-pytorch-sagemaker}{\texttt{https://\allowbreak{}github.com/\allowbreak{}philschmid/\allowbreak{}knowledge-\allowbreak{}distillation-\allowbreak{}transformers-\allowbreak{}pytorch-\allowbreak{}sagemaker}} to modify the original \texttt{Trainer} class from the HuggingFace Transformers library. Pretraining a 58M-parameter model with two teachers for 6 epochs takes less than 3 hours on a single NVIDIA RTX 3090. Training GPT-705M for 6 epochs takes around 12 hours, while training Llama-360M for 4 epochs takes around 2~hours.

\section{Dataset}
\label{sec:dataset}

The ``train'' dataset used in the \texttt{strict-small} track consists of approximately 10M words (as counted by the UNIX \texttt{wc} tool) that form a \emph{developmentally plausible} corpus, i.e.\ the sort of ``input'' that a typical child has access to: mostly transcribed speech and children's books. A separate, similar ``dev'' dataset of approximately 9.4M words is used for validation and testing. The entire dataset is in English, with some occasional foreign words such as e.g.\ proper nouns in Wikipedia articles.

Some simple, regex-based cleaning is performed on both datasets, e.g.\ to remove HTML tags from Wikipedia articles, non-verbal cues from subtitles, or even to correct I's that were incorrectly recognized as l's in OCR'ed uppercase text. The Python script responsible for the cleaning, \texttt{mrclean.py}, is included along with the model; it contains one function for each data source.

The cleaned dataset is then tokenized using Byte-Pair Encoding (BPE) with a vocabulary size of 16000. To avoid leakage, the tokenizer was trained \emph{exclusively} on the training split.
All the tokens are finally concatenated into a single one-dimensional vector.

Each split is divided into contiguous chunks of 128~tokens.
During each epoch of pretraining, the model is presented with a new random permutation of the chunks from the training split.\footnote{We noticed that adding a random offset between $0$ and $127$ to each chunk lead to marginally better performance; however, due to lack of time, the final teacher and student models were trained without such an offset.}
The validation loss is computed at the end of each epoch, by iterating in order over a fixed (but randomly sampled at the beginning) subset of the ``dev'' split.

\section{Performance}
\label{sec:performance}

\begin{figure*}
    \centering
    \includegraphics[width=0.92\textwidth]{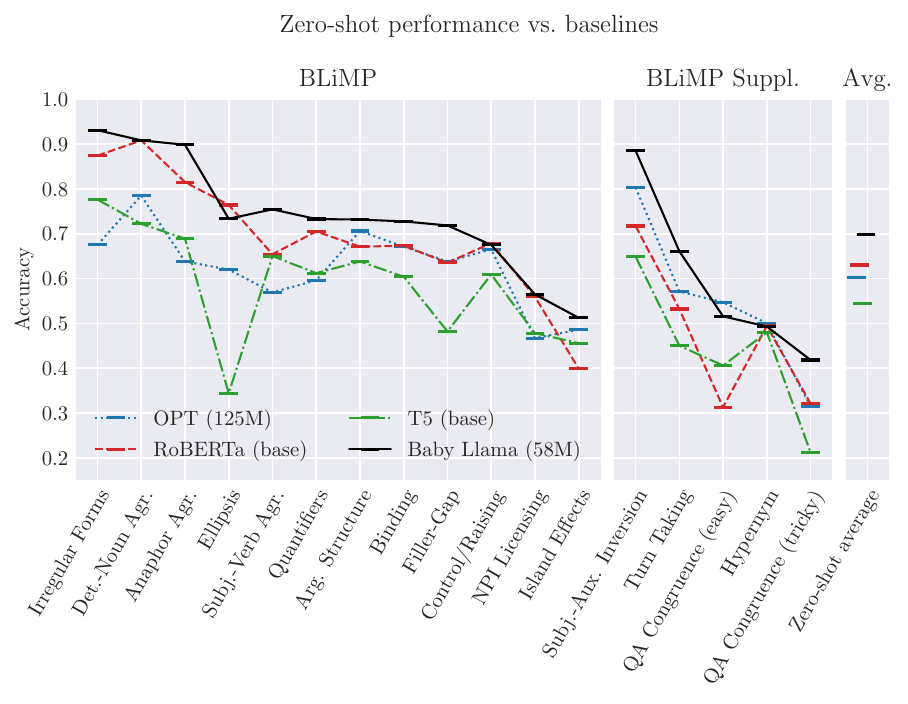}
    \caption{Parallel-coordinates plot summarizing the zero-shot performance of Baby Llama on the BLiMP and BLiMP Supplement benchmarks, compared with a number of baseline models.}
    \label{fig:zeroshot-baselines}
\end{figure*}

\begin{figure*}
    \centering
    \includegraphics[width=0.92\textwidth]{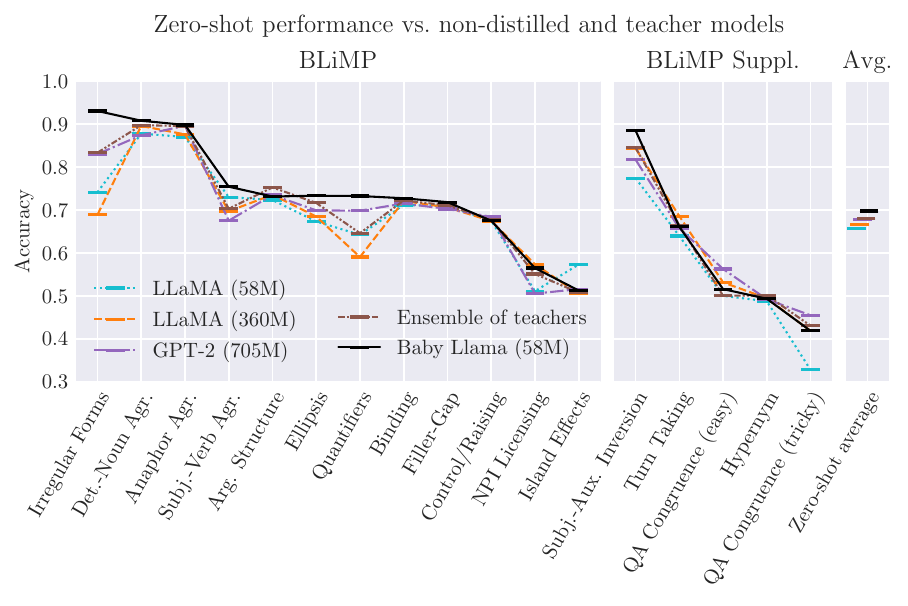}
    \caption{Parallel-coordinates plot summarizing the zero-shot performance of Baby Llama on the BLiMP and BLiMP Supplement benchmarks, compared with the same, non-distilled model, and both teacher models.}
    \label{fig:zeroshot-distil}
\end{figure*}

\begin{figure}
    \centering
    \includegraphics[width=\linewidth]{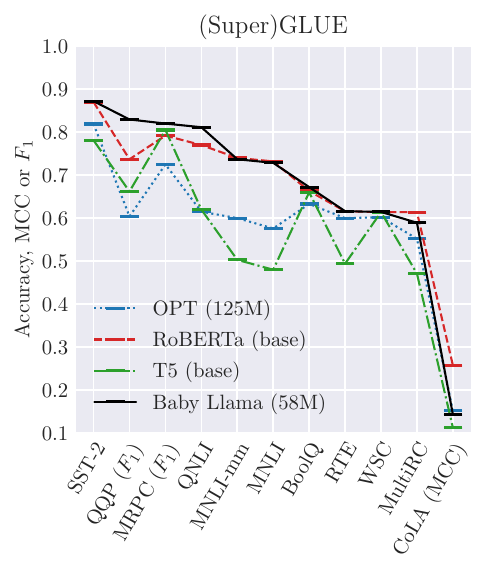}
    \caption{Parallel-coordinates plot summarizing the fine-tuning performance of Baby Llama on a subset of the GLUE and SuperGLUE benchmarks, compared with a number of baseline models. Unless specified otherwise, the metric used is the classification accuracy. The fine-tuning hyperparameters are listed in \cref{tab:fine-tuning-hyperparameters}.}
    \label{fig:superglue}
\end{figure}

\begin{figure}
    \centering
    \includegraphics[width=\linewidth]{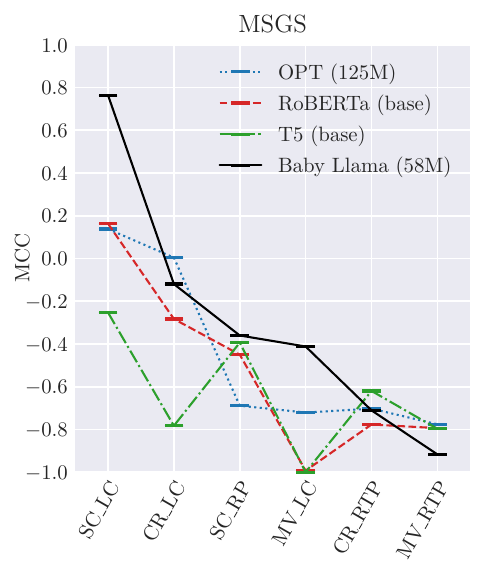}
    \caption{Parallel-coordinates plot summarizing the fine-tuning performance of Baby Llama on the MSGS benchmark, compared with a number of baseline models. The fine-tuning hyperparameters are listed in \cref{tab:fine-tuning-hyperparameters}.}
    \label{fig:msgs}
\end{figure}

Baby Llama is evaluated using a suite of linguistic benchmarks consisting of the BLiMP \citep{warstadt-etal-2020-blimp-benchmark} zero-shot benchmark (plus some yet-unpublished supplementary evals) as well as two fine-tuning benchmarks: SuperGLUE \citep{wang2020superglue} and MSGS \citep{warstadt-etal-2020-learning}. In \cref{app:evals}, we also discuss the model performance when used as part of an age-of-acquisition prediction task \citep{portelance2023predicting}.
These benchmarks are all run using the \texttt{lm-evaluation-harness} package \citep{eval-harness}, version \texttt{v0.2.0}.

We compare Baby Llama with three baseline models that are similar or larger in size and inference/fine-tuning computational cost: OPT (125M-parameter version, \citealp{zhang2022opt}), RoBERTa (base, 125M parameters, \citealp{liu2019roberta}) and T5 (base, 222M parameters, \citealp{raffel2020exploring}). The baseline models have been re-trained on the same 10M-word dataset by the organizers of the BabyLM challenge.\footnote{The checkpoints for those baseline models can be found at \url{https://huggingface.co/babylm}\,.}
For the BLiMP zero-shot benchmark, we add to the comparison the larger GPT-2 (705M) and LLaMA (360M) models that were used as teachers in the distillation procedure, a LLaMA (58M) model trained without distillation, as well as the ensemble model formed by averaging the output logits of both teachers. However, we do not evaluate the fine-tuning performance of these models due to the computational cost that it would incur.

The accuracy\footnote{We note that, despite seeding all the random number generators, we were not able to reproduce the numerical results across different machines (possibly due to different software versions) despite following as closely as possible the official procedure to install the evaluation pipeline. For consistency, all the reported results have been produced on the same machine.} of Baby Llama on the zero-shot benchmarks is presented in \cref{fig:zeroshot-baselines} along with the accuracy of the baselines, and in \cref{fig:zeroshot-distil} with that of the non-distilled and teacher models.
Its fine-tuning accuracy\footnote{Or MCC or $F_1$ score when explicitly mentioned.} is reported in \cref{fig:superglue} for (Super)GLUE, and its Matthews correlation coefficient (MCC) in \cref{fig:msgs} for MSGS. The performance is reported in the form of parallel-coordinates plots, with the lines serving as visual guides.
The full numerical results of the evals are listed in \cref{tab:blimp,tab:fine-tuning} in \cref{app:evals}.
% TODO: uncomment the sentence below in the preprint
They can also be found in a JSON file attached to the \LaTeX{} source of the present preprint.
%machine-readable format in Baby Llama's HuggingFace repository.

Baby Llama's performance is generally superior to all three baselines, for both zero-shot and fine-tuning benchmarks. It only falls significantly behind \emph{any} of the baselines on a handful of evals, thus showing a well-balanced and consistent overall performance.

Interestingly, Baby Llama not only performs better that both of the individual teacher models (as well as the non-distilled model) on most zero-shot tasks; it also performs better than the corresponding ensemble model. This clearly shows that the distillation procedure, by itself, leads to an improvement in the zero-shot accuracy.

When evaluating Baby Llama on the benchmarks that require fine-tuning, we noticed that the default fine-tuning hyperparameters suggested by the organizers lead to severe overfitting in a number of benchmarks (as evidenced by an increasing eval loss and no improvement --- or a decrease --- in the accuracy, while the training loss kept decreasing). To avoid this issue, we have re-tuned the fine-tuning hyperparameters as needed. The selected sets of hyperparameters are listed in \cref{tab:fine-tuning-hyperparameters}. For a small number of benchmarks, the performance didn't evolve smoothly as a function of the hyperparameters. Since this is symptomatic of overfitting on the eval dataset (making any comparison potentially inaccurate), we explicitly identify those benchmarks with the \textdagger{} symbol in \cref{tab:fine-tuning}.

\section{Conclusion}
\label{sec:conclusion}

In this work, we trained Baby Llama --- a 58M-parameter model based on the LLaMA architecture --- on the 10M-word BabyLM dataset using knowledge distillation.
It was distilled from an ensemble of two, inhomogeneous teachers: a 360M-parameter LLaMA model and a 705M-parameter GPT-2 model, both trained on the same dataset.
We observed that the model pretrained with the distillation loss \eqref{eq:distil-loss} performs better that the similar 58M-parameter model trained in the usual way. Moreover, the smaller, distilled model outperforms both of its teachers individually, as well as the ensemble model formed by the two teachers.

If those findings continue to hold at scale (see Limitations), they could help improve the sample efficiency of large language models, while reducing the amount of memory and compute necessary to deploy them. The increased sample efficiency could allow training larger, higher-performing models on the already-available training corpora (but at a higher training cost). Alternatively, it could limit the data collection necessary to train today's state-of-the-art models. This would e.g.\ allow focusing on higher-quality data, and it could be particularly useful in a hypothetical scenario where data collection gets restricted by online platforms, regulations, or due to copyright. Finally, the reduced size and computing requirements of the distilled model would reduce its energy footprint and facilitate on-device/local processing, leading to potentially-improved user privacy.

\section*{Limitations}

The results presented in this article have been obtained for models which are $10^3$ to $10^4$ times smaller that current state-of-the-art language models. Many important properties of these models have been shown to emerge as the model size increases \citep{gpt2,gpt3}. Therefore, the results obtained at small scales may not necessarily generalize to larger scales.

Furthermore, our results have been obtained in the regime where the number of parameters significantly exceeds the number of training tokens. This differs from today's state-of-the-art language models, which are usually trained on many more tokens than their number of parameters, e.g.\ $\sim 20$ times more for models trained in a compute-optimal way following \citet{hoffmann2022training}. Such models may not have the luxury to dedicate as many parameters to a given piece of information or feature as ours. Therefore, there is no guarantee that the nearly lossless distillation that we have observed will generalize to such models.

Due to these differences in scale and tokens-to-parameters ratio, it is not clear if our proposed distillation procedure could be scaled up in order to increase the sample efficiency of today's largest language models.
Although this hypothesis can in principle be tested experimentally, the authors lack the computational resources required to perform such a test.

Finally, our results have been obtained for a textual training corpus, in the context of language modeling. Further experimentation will be required in order to investigate whether our findings generalize to different data modalities and to other domains where transformer-based models are also being used.

\section*{Acknowledgements}

We thank Oleg Ruchayskiy and Troels C.\ Petersen for their support.
JLT acknowledges partial financial support by the Spanish Research Agency (Agencia Estatal de Investigaci\'on) through the grant IFT Centro de Excelencia Severo Ochoa No CEX2020-001007-S, by the grant PID2019-108892RB-I00 funded by MCIN/AEI/ 10.13039/501100011033, by the European Union's Horizon 2020 research and innovation programme under the Marie Sk\l odowska-Curie grant agreements No 860881-HIDDeN and Staff Exchange grant agreement No 101086085 - ASYMMETRY, and by the grant Juan de la Cierva FJC2021-047666-I funded by MCIN/AEI/10.13039/501100011033 and by the European Union ``NextGenerationEU''/PRTR.
The work of IT was partially supported by the Carlsberg foundation, and by
the European Union's Horizon 2020 research and innovation program under the Marie Sklodowska-Curie grant agreement No. 847523 `INTERACTIONS'.

\bibliography{anthology,custom}
\bibliographystyle{acl_natbib}

\appendix

\section{Evals}
\label{app:evals}

The numerical results for the zero-shot accuracy on the BLiMP suite of benchmarks \citep{warstadt-etal-2020-blimp-benchmark} can be found in \cref{tab:blimp}, while the results for the fine-tuning accuracy on the SuperGLUE \citep{wang2020superglue} and MSGS \citep{warstadt-etal-2020-learning} benchmarks are listed in \cref{tab:fine-tuning}.
Finally, the hyperparameters selected for the various fine-tuning tasks are summarized in \cref{tab:fine-tuning-hyperparameters}.

\paragraph{Age-of-acquisition prediction}
In addition to the above-mentioned benchmarks, we have tested our model on an age-of-acquisition task proposed by \citet{portelance2023predicting}.
Its aim is to predict the median age at which a word is learned by children, as a function of a number of variables (such as the lexical category, concreteness, frequency, etc.), using a \emph{linear model}.
One of these variables is the average \textit{surprisal}, i.e.\ the average negative log-probability of the word across all the contexts where it appears, as predicted by a causal language model. This is the only place where the language model enters.
The use of this task as a benchmark for language models fundamentally relies on the assumption of a linear relationship between the surprisal and the age of acquisition. If this assumption is true, then a more accurate estimation of the tokens probabilities by the language model should indeed translate into a more accurate prediction of the age of acquisition. If, however, this assumption is not justified, then the linear model --- but not the language model --- might be the bottleneck, and a better language model won't necessarily lead to a better prediction.

The mean absolute deviations of the predicted ages of acquisition are reported for various language models and lexical categories in \cref{tab:aoa}.
We can only observe minor differences between the four considered language models (likely due to random noise), suggesting that the linear regression --- and not the language model --- is indeed the bottleneck.
Therefore, this task is unlikely to be indicative of the performance of Baby Llama relative to the baselines.

\begin{table*}
\centerline{\small%
\begin{tabular}{llcccccccc}
\toprule
& Model & \makecell{OPT \\ (125M)} & \makecell{RoBERTa \\ (base)} & \makecell{T5 \\ (base)} & \makecell{LLaMA \\ (58M)} & \makecell{LLaMA \\ (360M)} & \makecell{GPT-2 \\ (705M)} & \makecell{Ensemble \\ of teachers} & \makecell{Baby Llama \\ (58M, distilled)} \\
\midrule
\multirow{12}{*}{\rotatebox{90}{BLiMP}} & Anaphor Agr. & 63.8 & 81.5 & 68.9 & 87.0 & 87.6 & 89.6 & 89.6 & \textbf{89.8} \\
& Arg. Structure & 70.6 & 67.1 & 63.8 & 72.3 & 73.5 & 73.5 & \textbf{75.3} & 73.1 \\
& Binding & 67.1 & 67.3 & 60.4 & 71.2 & 72.1 & 71.5 & 72.2 & \textbf{72.7} \\
& Control/Raising & 66.5 & 67.9 & 60.9 & 67.5 & 67.4 & \textbf{68.4} & 67.7 & 67.5 \\
& Det.-Noun Agr. & 78.5 & \textbf{90.8} & 72.2 & 87.8 & 89.6 & 87.4 & 89.8 & \textbf{90.8} \\
& Ellipsis & 62.0 & \textbf{76.4} & 34.4 & 67.3 & 68.5 & 69.9 & 71.7 & 73.3 \\
& Filler-Gap & 63.8 & 63.5 & 48.2 & 70.9 & 70.6 & 70.2 & 71.1 & \textbf{71.8} \\
& Irregular Forms & 67.5 & 87.4 & 77.6 & 74.1 & 68.9 & 83.1 & 83.4 & \textbf{93.1} \\
& Island Effects & 48.6 & 39.9 & 45.6 & \textbf{57.3} & 50.4 & 51.6 & 50.7 & 51.2 \\
& NPI Licensing & 46.7 & 55.9 & 47.8 & 51.1 & \textbf{57.3} & 50.5 & 55.1 & 56.5 \\
& Quantifiers & 59.6 & 70.5 & 61.2 & 64.2 & 59.0 & 69.8 & 64.6 & \textbf{73.3} \\
& Subj.-Verb Agr. & 56.9 & 65.4 & 65.0 & 73.0 & 69.7 & 67.5 & 70.3 & \textbf{75.4} \\
\midrule
\multirow{5}{*}{\rotatebox{90}{BLiMP suppl.}} & Hypernym & 50.0 & 49.4 & 48.0 & 48.7 & 49.4 & 49.2 & \textbf{50.1} & 49.3 \\
& QA Congruence (easy) & 54.7 & 31.3 & 40.6 & 50.0 & 53.1 & \textbf{56.2} & 50.0 & 51.6 \\
& QA Congruence (tricky) & 31.5 & 32.1 & 21.2 & 32.7 & 41.8 & \textbf{45.5} & 43.0 & 41.8 \\
& Subj.-Aux. Inversion & 80.3 & 71.7 & 64.9 & 77.4 & 84.3 & 81.7 & 84.5 & \textbf{88.5} \\
& Turn Taking & 57.1 & 53.2 & 45.0 & 63.9 & \textbf{68.6} & 65.7 & 66.4 & 66.1 \\
\bottomrule
\end{tabular}}
\caption{Zero-shot accuracy (in percent), as evaluated by the BLiMP suite of benchmarks (top) and some supplementary benchmarks (bottom).}
\label{tab:blimp}
\end{table*}

\begin{table*}
    \centering
    \begin{tabular}{llcccc}
    \toprule
    & Model & \makecell{OPT \\ (125M)} & \makecell{RoBERTa \\ (base)} & \makecell{T5 \\ (base)} & \makecell{Baby Llama \\ (58M, distilled)} \\
    \midrule
    \multirow{11}{*}{\rotatebox{90}{(Super)GLUE}} & CoLA (MCC) & 15.2 & \textbf{25.8} & 11.3 & 14.3 \\
    & SST-2 & 81.9 & 87.0 & 78.1 & \textbf{87.2} \\
    & MRPC ($F_1$) & 72.5 & 79.2 & 80.5 & \textbf{82.0} \\
    & QQP ($F_1$) & 60.4 & 73.7 & 66.2 & \textbf{83.0} \\
    & MNLI & 57.6 & \textbf{73.2} & 48.0 & 72.9 \\
    & MNLI-mm & 60.0 & \textbf{74.0} & 50.3 & 73.7 \\
    & QNLI & 61.5 & 77.0 & 62.0 & \textbf{81.1} \\
    & RTE & 60.0 & \textbf{61.6} & 49.4 & \textbf{61.6}\textsuperscript{\textdagger} \\
    & BoolQ & 63.3 & 66.3 & 66.0 & \textbf{67.2}\textsuperscript{\textdagger} \\
    & MultiRC & 55.2 & \textbf{61.4} & 47.1 & 58.9\textsuperscript{\textdagger} \\
    & WSC & 60.2 & \textbf{61.4} & \textbf{61.4} & \textbf{61.4}\textsuperscript{\textdagger} \\
    \midrule
    \multirow{6}{*}{\rotatebox{90}{MSGS (MCC)}} & CR\_LC & \textbf{0.4} & -28.3 & -78.3 & -12.0 \\
    & CR\_RTP & -70.3 & -77.7 & \textbf{-62.0} & -71.1 \\
    & MV\_LC & -72.1 & -99.3 & -100.0 & \textbf{-41.2} \\
    & MV\_RTP & \textbf{-77.6} & -79.4 & -79.7 & -91.7 \\
    & SC\_LC & 13.8 & 16.3 & -25.3 & \textbf{76.4} \\
    & SC\_RP & -68.9 & -45.0 & -39.4 & \textbf{-36.0} \\
    \bottomrule
    \end{tabular}
    \caption{Fine-tuning accuracy (if not specified), Matthews correlation coefficient (MCC) or $F_1$ score --- in percent --- as evaluated by the SuperGLUE (top) and MSGS (bottom) suites of benchmarks. The \textdagger{} symbol indicates benchmarks for which the best performance was reached only for a narrow range of hyperparameters, suggesting possible overfitting of the validation set.}
    \label{tab:fine-tuning}
\end{table*}

\begin{table*}
    \centering
    \begin{tabular}{llcccccc}
    \toprule
    & Task & Max.\ learning rate & Batch size & Max.\ epochs & Patience & Eval every & Seed \\
    \midrule
    \multirow{11}{*}{\rotatebox{90}{(Super)GLUE}} & CoLA & $4\cdot 10^{-5}$ & 64 & 3 & 10 & 20 & 12 \\
    & SST-2 & $5\cdot 10^{-5}$ & 64 & 6 & 10 & 200 & 12 \\
    & MRPC & $3\cdot 10^{-5}$ & 64 & 3 & 10 & 20 & 12 \\
    & QQP & $4\cdot 10^{-5}$ & 64 & 10 & 10 & 1000 & 12 \\
    & MNLI & $5\cdot 10^{-5}$ & 64 & 6 & 10 & 200 & 12 \\
    & MNLI-mm &$5\cdot 10^{-5}$ & 64 & 6 & 10 & 200 & 12 \\
    & QNLI & $5\cdot 10^{-5}$ & 64 & 6 & 10 & 200 & 12 \\
    & RTE & $5\cdot 10^{-5}$ & 64 & 6 & 10 & 200 & 12 \\
    & BoolQ & $3\cdot 10^{-4}$ & 16 & 10 & 10 & 10 & 12 \\
    & MultiRC & $1\cdot 10^{-4}$ & 64 & 7 & 10 & 1000 & 42 \\
    & WSC & $5\cdot 10^{-7}$ & 1 & 10 & 1000 & 2000 & 12 \\
    \midrule
    \multirow{6}{*}{\rotatebox{90}{MSGS}} & CR\_LC & $1\cdot 10^{-3}$ & 64 & 2 & 10 & 10 & 12 \\
    & CR\_RTP & $5\cdot 10^{-5}$ & 64 & 6 & 10 & 200 & 12 \\
    & MV\_LC & $5\cdot 10^{-5}$ & 64 & 6 & 10 & 200 & 12 \\
    & MV\_RTP & $5\cdot 10^{-5}$ & 64 & 6 & 10 & 200 & 12 \\
    & SC\_LC & $1\cdot 10^{-3}$ & 64 & 2 & 10 & 10 & 12 \\
    & SC\_RP & $1\cdot 10^{-3}$ & 64 & 2 & 10 & 10 & 12 \\
    \bottomrule
    \end{tabular}
    \caption{List of the hyperparameters selected when fine-tuning Baby Llama on the various evals that require fine-tuning.}
    \label{tab:fine-tuning-hyperparameters}
\end{table*}

\begin{table*}
    \centering
    \begin{tabular}{lcccc}
    \toprule
    Model & \makecell{OPT \\ (125M)} & \makecell{RoBERTa \\ (base)} & \makecell{T5 \\ (base)} & \makecell{Baby Llama \\ (58M, distilled)} \\
    \midrule
    AoA Overall (591 words) & \textbf{2.03} & 2.06 & 2.04 & 2.06\\
    AoA Nouns (322 words) & 1.98 & 1.99 & \textbf{1.97} & 1.99 \\
    AoA Predicates (167 words) & \textbf{1.81} & 1.85 & 1.82 & 1.84 \\
    AoA Function words (102 words) & \textbf{2.57} & 2.65 & 2.64 & 2.63 \\
    \bottomrule
    \end{tabular}
    \caption{Performance of the model, when used as part of an age-of-acquisition (AoA) prediction task for various lexical categories, as quantified by the mean absolute deviation (lower is better).}
    \label{tab:aoa}
\end{table*}

\section{Other attempts and null results}
\label{app:attempts}
In this appendix, we briefly describe various approaches that we have investigated in order to improve the performance of our models. Unlike distillation from an ensemble of teachers, those attempts had mixed results and we haven't pursued them further, in part due to our limited computational resources.

\paragraph{Curriculum learning}
We implemented a simple version of curriculum learning, directly inspired by the original paper from \citet{curriculum-learning}. We split the 10 files composing the training set into 5 buckets, in order of roughly increasing complexity according to some readability metrics\footnote{The metrics used are the Flesch reading ease, Flesch-Kincaid grade level, Gunning fog index, automated readability index, and SMOG grade. The buckets are 1.\ \texttt{aochildes}, 2.\ \texttt{open\_subtitles}, 3.\ \texttt{switchboard}, \texttt{cbt}, \texttt{qed}, \texttt{children\_stories}, \texttt{bnc\_spoken}, 4.\ \texttt{simple\_wikipedia}, \texttt{gutenberg} and 5.\ \texttt{wikipedia}.} computed using the \texttt{textstat} Python package. We start training for 3~epochs using the lowest bucket only, then, every 3~epochs, we add the next bucket to the training set without removing the previous ones, until we have trained for 3~epochs on the full training set. The full validation set is always used to compute the eval loss.

After training a 10M-parameter GPT-2 model using the schedule described above, the eval loss\footnote{In this experiment and the others described in this appendix, the exact tokenizer and sequence length may differ from the ones used for the final Baby Llama, therefore the loss isn't directly comparable.} plateaued at 3.75, comparable to the 3.74 obtained by training the same model for the same wall-clock duration but using the full training set from the beginning. Although the model trained with curriculum learning scored on average 1 percentage point above the non-curriculum model on the zero-shot benchmarks, the overall picture was mixed, due to significant regressions in two of the evals.
The absence of a significant improvement from curriculum learning is in line with previously-reported negative results in \citet{surkov-etal-2022-data}, although we should remain cautious since our attempt wasn't comprehensive and modern sampling methods may lead to significantly better results.

\paragraph{Switch Transformer}
Using the HuggingFace Transformers library, we have implemented a decoder Switch Transformer \citep{switch-transformers} for causal language modeling, based on the encoder-decoder version available in said library. This mixture-of-experts model was initially introduced to scale up the number of parameters at a constant computational cost.

We train both a GPT2-10M baseline\footnote{8 layers, embedding dimension 256, 16 heads, and vocabulary size 16000.}, as well as a number of Switch Transformers with the same number of layers and embedding dimension but different numbers of experts and expert capacities (tuning separately the other hyperparameters of each model).
We observe, as expected, that a Switch Transformer with a single expert of capacity~1 closely matches the performance of the baseline GPT-2 model.
However, as we scale up the number of experts and expert capacity, we observe a performance degradation (both in the loss and zero-shot scores), even after allowing for longer training of the larger models.
This suggests that mixture-of-experts models may not bring any advantages for the model and dataset sizes considered here.

\paragraph{Ensembling of homogeneous models}
We averaged the predicted logits of 4 GPT2-10M models trained from different random initializations, but otherwise identical, and compared the results of the ensemble with those of its constituent GPT2-10M models. All models had their hyperparameters tuned to minimize the eval loss. While the individual models had an average eval loss of 3.77, the averaged model reached 3.66, an improvement of 0.11. This translates into an improvement of 1 to 2 percentage points (depending on the specific seed) in the average zero-shot BLiMP score; more importantly, the averaged model always scored higher than the average score of its constituents, and often scored higher than all of them.
Despite the initial success of this method, adding more teachers to the ensemble from which Baby Llama was distilled yielded no further improvement to the performance of the distilled model, suggesting that the gains from using this method do not sum with those from knowledge distillation.

\paragraph{Sharpness-Aware Minimization}
\citet{2020arXiv201001412F} have introduced a Sharpness-Aware Minimization (SAM) procedure for simultaneously minimizing the loss value and its sharpness. It has been shown in \citet{2021arXiv211008529B} that applying SAM when fine-tuning on multiple downstream tasks can result in substantial performance gains. 

Here we tried a rather different approach: we pretrained GPT2-10M in the usual way for 15 epochs and then trained it using SAM for one more epoch. This approach in some sense resembles the gradient ascent discussed in the next paragraph.
We implemented a custom training loop for HuggingFace Transformers models based on \url{https://github.com/karpathy/nanoGPT}.
This allowed us to use the two-step SAM optimization from \url{https://github.com/davda54/sam}.
Unfortunately, we have not observed any improvement in the model's zero-shot capabilities resulting from this type of SAM application.

\paragraph{Post-training Gradient Ascent}
\citet{2023arXiv230607052Y} have empirically demonstrated that a few steps of Gradient Ascent Post-training (GAP) enhances the zero-shot generalization capabilities across diverse NLP tasks. 

In order to test GAP, we first applied it to GPT2-10M. We took a fully trained model and performed 15 to 100 steps of gradient ascent (following the original paper, we used batch size 1 and learning rate $5\cdot 10^{-5}$). We observed some improvements on BLiMP (although those were not consistent among the various tasks). However, we did not manage to further improve the zero-shot performance of the distilled Baby Llama, suggesting again that the gains from using this method do not sum with those from knowledge distillation.

\end{document}